\pdfoutput=1

\documentclass[11pt]{article}

\usepackage{acl2023}

\usepackage{times}
\usepackage{latexsym}

\usepackage[T1]{fontenc}

\usepackage[utf8]{inputenc}

\usepackage{microtype}

\usepackage{inconsolata}

\usepackage{multirow}
\usepackage{subfigure}
\usepackage{booktabs}
\usepackage{graphicx}
\usepackage{algorithm}
\usepackage{algorithmic}
\usepackage{amsmath}
\title{Improve Meta-learning for Few-Shot Text Classification with All You Can Acquire from the Tasks}

\author{Xinyue Liu, Yunlong Gao, Linlin Zong, Bo Xu\thanks{*Corresponding author} \\
        Dalian University of Technology, Dalian, China\\
        xyliu@dlut.edu.cn, yunlong@mail.dlut.edu.cn, llzong@dlut.edu.cn, xubo@dlut.edu.cn  \ }

\begin{document}
\maketitle
\begin{abstract}

Meta-learning has emerged as a prominent technology for few-shot text classification and has achieved promising performance. However, existing methods often encounter difficulties in drawing accurate class prototypes from support set samples, primarily due to probable large intra-class differences and small inter-class differences within the task. Recent approaches attempt to incorporate external knowledge or pre-trained language models to augment data, but this requires additional resources and thus does not suit many few-shot scenarios. In this paper, we propose a novel solution to address this issue by adequately leveraging the information within the task itself. Specifically, we utilize label information to construct a task-adaptive metric space, thereby adaptively reducing the intra-class differences and magnifying the inter-class differences. We further employ the optimal transport technique to estimate class prototypes with query set samples together, mitigating the problem of inaccurate and ambiguous support set samples caused by large intra-class differences. We conduct extensive experiments on eight benchmark datasets, and our approach shows obvious advantages over state-of-the-art models across all the tasks on all the datasets. For reproducibility, all the datasets and codes are available at https://github.com/YvoGao/LAQDA.

\end{abstract}

\section{Introduction}

\begin{figure*}
\centering
\includegraphics[width=0.9\textwidth]{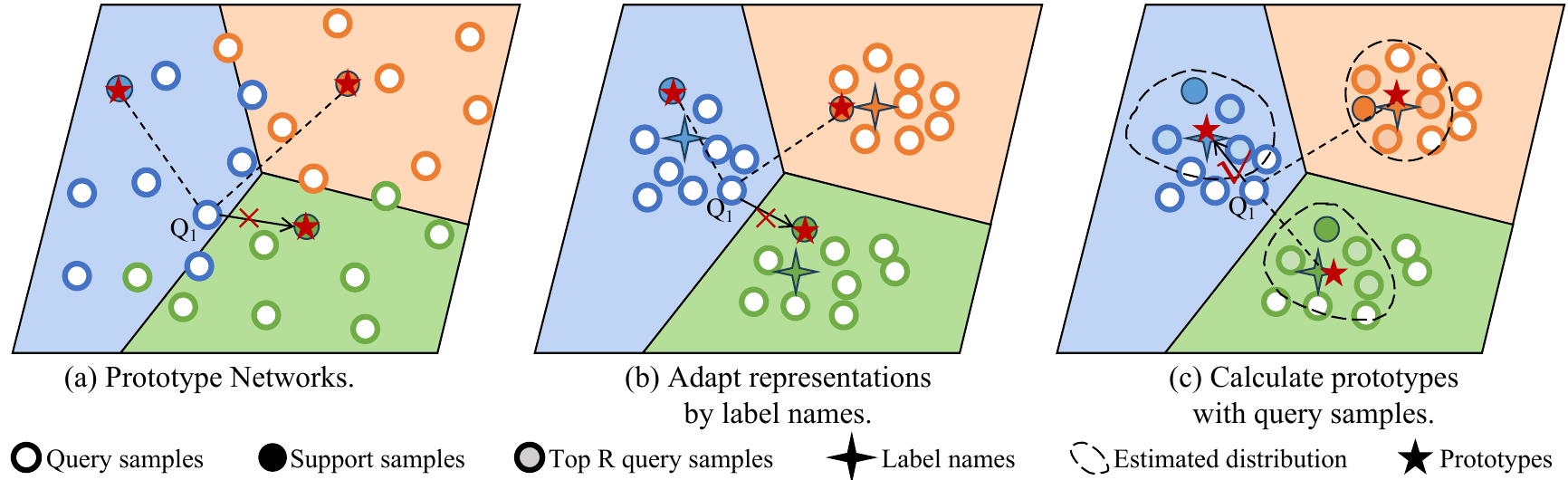} 
\caption{The feature space illustration of a 3-way 1-shot task. Figure (a) shows that Prototypical Networks (PN) classifies query samples by prototypes calculated by the support set. Since the given query sample $Q_1$ whose true label is \textit{blue} is the closest to the estimated prototype of the \textit{green} class, PN misclassifies $Q_1$ to \textit{green} class. Figure (b) shows that sample representations are closer adapted by label names but the given query sample $Q_1$ is still misclassified to \textit{green} class due to the intra-class differences. Figure (c) shows the classifier calculates prototypes from support and query samples, and $Q_1$ is classified to \textit{blue} class correctly.}
\label{fig:idea}
\end{figure*}

Text classification is a fundamental task in the natural language processing (NLP) theme, which has been widely applied to various real applications. However, a deficiency of supervised data is often experienced in the real world. Few-shot text classification \citep{Yu_2018, Geng2019} aims to detect novel categories with very limited labeled examples by using knowledge learned from known categories, which is crucial for many applications but remains to be a challenging task.

The existing methods can be broadly categorized into two main branches. One branch is Transfer learning \citep{gpt_2020, Gupta_2021, proto_22}, which aims to leverage general-domain knowledge acquired from Pre-trained Language Models (PLMs). While prompt learning techniques have shown superior results in transfer learning, these methods often require large-scale language models (LLMs) and are more suitable for explicit and simplistic classification tasks such as emotion recognition (positive or negative). As a result, they may not be applicable in many real-world scenarios, particularly when computing resources are limited, such as those in mobile devices. And prompting often needs much manual work and is also some kind of supervision, thus it does not suit many few-shot scenarios.

The other branch is Meta-learning \citep{proto_2017, Bao_2019, Luo_2021}, which aims to learn cross-task transferable knowledge rather than recalling pre-trained knowledge gained through PLMs. These methods employ small-scale models and do not have a bias towards specific target problems, and are more suitable for practical applications in few-shot scenarios. Typical meta-learning methods, e.g., Prototypical Networks (PN) \citep{proto_2017}, which leverages Euclidean distance to measure query examples against the class vector averaged by support examples, often meets with an overfitting issue. As illustrated in Figure \ref{fig:idea}(a), the query sample $Q_1$, which belongs to the \textit{blue} class, is erroneously classified as \textit{green} class because it is closer to the estimated prototype of the \textit{green} class.

To alleviate the overfitting issue of PN, RPOTAUG~\citep{Dopierre_Gravier_Logerais_2021} and MEDA~\citep{MEDA} leverage data augmentation to expansion support set. MLADA~\citep{Han__2021} introduces an adversarial domain adaptation network for reducing intra-class differences. ContrastNet~\citep{Chen_Zhang_Mao_Xue_2022} magnifies the inter-class differences by contrastive learning. However, the overfitting issue still exists because of the randomness of the sampled support sets and probable large intra-class differences. As shown in Figure \ref{fig:idea}(b), despite the small intra-class difference and large inter-class difference, $Q_1$ is still misclassified as \textit{green} class because the support samples are located far away from the class centers, resulting in the estimated prototype of the \textit{green} class being closer to $Q_1$.

Although previous works have achieved certain improvements over PN, it is worth noting that most of them focus on designing complex structures or incorporating external data augmentation, overlooking the valuable knowledge presented within the task itself. In this paper, we propose a method called LAQDA to address the overfitting issue for few-shot text classification. We also use the PN framework, however, by introducing the \textbf{L}abel-\textbf{A}dapter and \textbf{Q}uery-\textbf{D}ata-\textbf{A}ugmenter modules, our method estimates class prototypes that are closer to the class centers. Specifically, we design a Label-Adapter module that constructs a task-adaptive metric space by attention mechanism, which clusters the same class sample representations, thereby adaptively reducing the intra-class differences and magnifying the inter-class differences. We further design a Query-Data-Augmenter module to estimate class prototypes with the query set samples together by the optimal transport technique, mitigating the problem of inaccurate and ambiguous support set samples caused by intra-class differences. For example, as shown in Figure \ref{fig:idea}(c), the estimated prototypes by LAQDA are closer to the class centers, and the query sample $Q_1$ is closest to the prototype of the \textit{blue} class, so it is classified correctly. We evaluate the proposed method on eight popular datasets for few-shot text classification, and our approach shows obvious advantages over state-of-the-art models across all the tasks on all the datasets.

\section{Our Method}

\subsection{Problem Formulation}
\label{sec:PF}

Meta-learning paradigm of few-shot text classification follows the \textbf{$N$-way $K$-shot task setting}, i.e., for each task, there are $N$ classes and each class has $K$ supports (labeled samples). Specifically, the data is divided into two parts: the source classes $Y_{train}$, target classes $Y_{test}$, and $Y_{train} \cap Y_{test}=\emptyset$. In general, meta-learning contains two phases: meta-training and meta-testing. 

\noindent \textbf{Meta-training} The model is trained with numerous tasks. For each task, $N$ classes are sampled from training data $Y_{train}$, $K$ labeled examples are sampled as the support set $S$ and another $M$ examples as the query set $Q$ per class, donated as $S = {(X_i, Y_i)}^{N \times K} _{i=1}$ and $Q = {(X_j, Y_j)}^{N \times M} _{j=1}$. The model makes predictions about the query set $Q$ based on the given support set $S$. Then the model updates the parameters by minimizing the loss in the query set $Q$. 

\noindent \textbf{Meta-testing} For each task, $N$ novel classes will be sampled from $Y_{test}$, which is disjoint to $Y_{train}$. Then the support set $S$ and the query set $Q$ will be sampled from the $N$ classes like in meta-training. The performance of the model will be evaluated through the average classification accuracy on the query set $Q$ across all the testing tasks.

\begin{figure*}[]
\centering
\includegraphics[width=.9\textwidth]{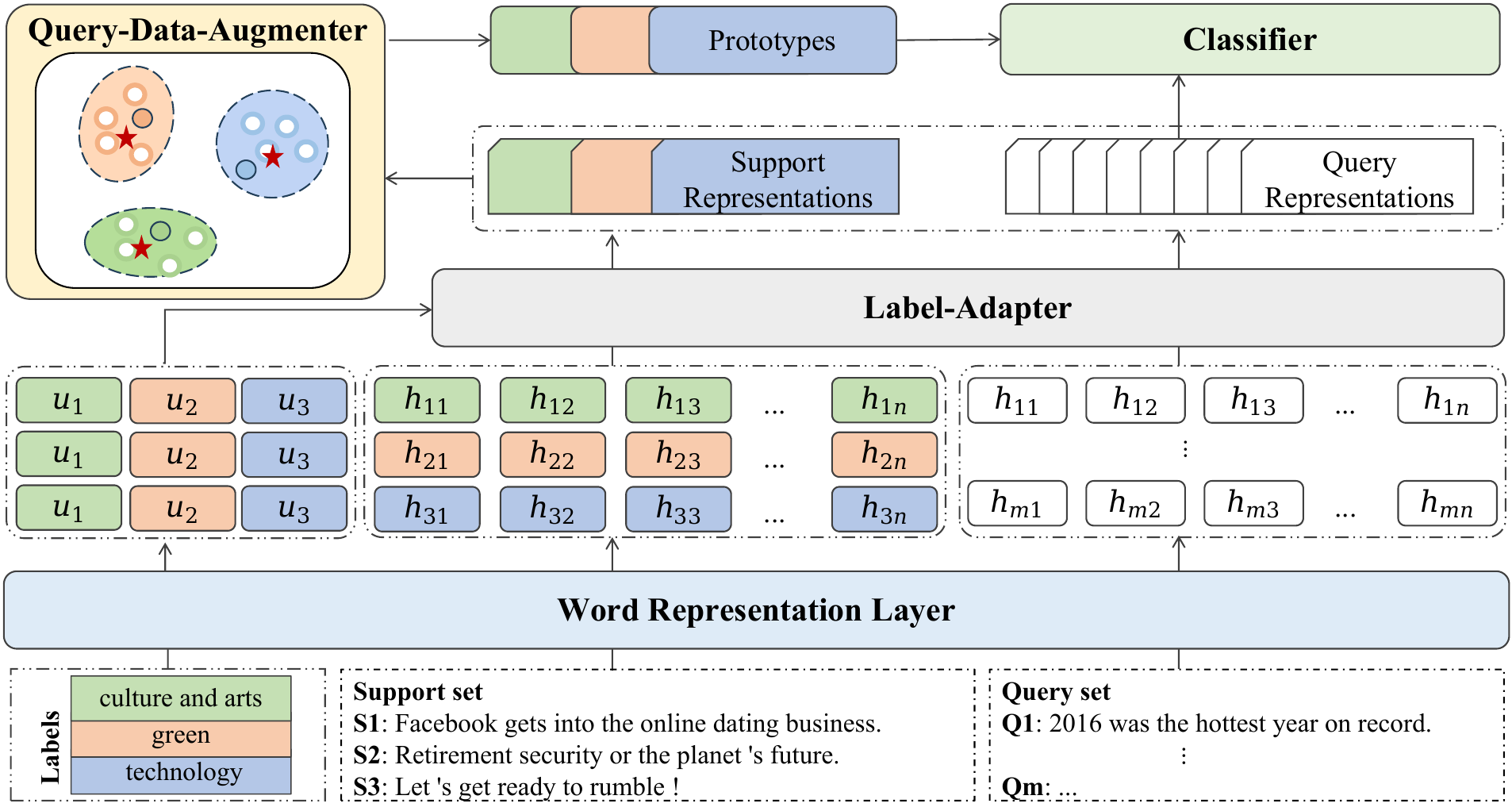} 
\caption{ The process of LAQDA on a 3-way 1-shot few-shot news classification task. \textit{Word Representation Layer} maps sentences and label names into $h$ and $u$. \textit{Label-Adapter} generates sample representations for the task. \textit{Query-Data-Augmenter} calculates prototypes with support set and query set. \textit{Classifier} outputs the final result.}
\label{fig:model}
\end{figure*}

\subsection{Framework}
\label{sec:Overall}

In our work, we resort to exploring the information within the task itself to boost the performance of few-shot text classification. In contrast to previous methods focusing on complex models or external knowledge, our method merely uses the label names and query samples of the task. Figure \ref{fig:model} gives an overview of our framework, which mainly consists of four modules. First, the \textit{Word Representation Layer} gets the word vector representations from the input sentences and label names. Second, the \textit{Label-Adapter} joints the label names and samples to generate the intra-class closer sample representations. Third, the \textit{Query-Data-Augmenter} leveraging query samples as data augmentation to calculate the class prototypes, mitigating the problem of inaccurate and ambiguous support set samples caused by probable large intra-class differences. Finally, query samples are inferred by a \textit{Classifier}. Note that the Word Representation Layer and the Classifier modules use the well-known word-embedding technique and the initial Prototypical Networks, next we mainly introduce the Label-Adapter and the Query-Data-Augmenter modules. 

\subsection{Label-Adapter}
\label{sec:LA}

To construct a task-adaptive metric space that generates closer intra-class sample representations and larger inter-class differences within a task, we encode the words of each sentence and task label names simultaneously. We use the self-attention layer as the building block of this module, due to its inherent weighting mechanism of pairwise similarities between elements in the sequence. Specifically, we first concatenate a learned prefix $h_0$, each sentence sequence $[h_1, h_2, ..., h_n]$, and the task label names $[u_1, u_2, ..., u_N]$. 
$h_0$ can be initialized by the mean of word vectors in each sentence.

We then adopt the set multi-head attention block, which plays the role of an adapter with trainable meta-parameters $\theta$, and is defined as:
\begin{equation}
    LA_\theta (Q, K, V ) = \sigma(QK^T ) \cdot V,
    \label{eq:la}
\end{equation}
where the pairwise dot-product $QK^T$ measures the similarity amongst features and is used for feature weighting computed through an activation function $\sigma$. Intuitively, each feature of $V$ will get more weight if the dot-product between $Q$ and $K$ is larger.

In Eq. \ref{eq:la}, following the self-attention mechanism, we have $Q=K=V$, the input is the sequence of $[h_0, h_1, ..., h_n, u_1, ..., u_N]$, and the output is a vector of learned parameters $h_0^*$ as a sample representation $v$, as expressed in:
\begin{equation}
    v = h_0^* = LA_\theta ([h_0, h_1, ..., h_n, u_1, ..., u_N]).
    \label{eq:rep}
\end{equation}

By representing each sentence with the corresponding output $v$, we get a new metric space in which the representation of each sentence is closer to the corresponding class center, and will be further used in the next module.

\subsection{Query-Data-Augmenter}
\label{sec:OT}

Due to the randomness of the sampled support sets and intra-class differences, the prototypes obtained from the support set $S$ may not be accurate and representative. While query set $Q$ contains abundant unlabeled samples belonging to $N$ classes, thus we estimate class prototypes utilizing the information of query samples that are top $R$ similar to per support set sample, as the more samples, the better prototypes can be calculated.

We use the Optimal Transport (OT) technique, which can help transfer data efficiently between discrete empirical distributions, to transfer query samples to support set to estimate class prototypes that are closer to the class centers by utilizing the query samples. 

Consider an $N$-way $K$-shot task, given a novel class $c$, its $K$ support samples denoted as $\{x^S_{c1}, ..., x^S_{cK}\}$, query samples $\{x^Q_1, ...,x^Q_m \}$ and their representations $\{v^S_{c1}, ..., v^S_{cK} \}$, $\{v^Q_1, ...,v^Q_m \}$, we treat each sample as a random variable which follows the Gaussian distribution. Specifically, for the $c$-th class support set $S_c$, we first retrieve its $R$ most similar samples in the query set $Q$ based on the OT distance,

\begin{equation}
\begin{aligned}
    M_c^Q & = \mathop{\arg \min} \limits_{c \in N} \mathcal{W} (Q, S_c) \\
    & = \arg \min \limits_{c \in N}  \min \limits_{\mathbf{T}\in \Sigma(Q,x^S_{ci})} <\mathbf{C}, \mathbf{T}> \\
    & = \{a_1, a_2, ..., a_R \},
\end{aligned}
\end{equation}
where $\mathbf{C}$ is a cost matrix with each element computed as: $c(x^Q_i, x^S_j ) =||v^Q_i-v^S_j ||^2_2$, $\mathbf{T} \in \mathbf{R}_{+}^{n \times m}: \{ \mathbf{T} 1_{m}= Q, \mathbf{T} 1_{n}= S_c\}$. We denote the augment information for the $c$-th class support set as $M^Q_c$, and the optimal transport plan between $S_c$ and $Q$ as $\mathbf{T}_c$, which could be obtained through the Sinkhorn algorithm \citep{OTal}.

We next adapt the augment information $M_c^Q$ from the query set $Q$, mapping to the task as follows:

\begin{equation}
\hat{a_i}=\arg \min \limits_{a_i \in M_c^Q} \sum_{j} \mathbf{T}_{c}(i, j) \cdot c \left(a_i,v^S_{cj}\right),
\end{equation}
for all $i = 1, . . . , R,$ where $\hat{a_i}$ denotes the projected representation of the $i$-th sample representation in $M_c^Q$ , and $\mathbf{T}_c(i, j)$ represents an element of the optimal transport plan $\mathbf{T}_c$. It has been shown that when the cost function is squared Euclidean norm, the solution to the above barycenter mapping corresponds to a weighted average of $S_c$ \citep{OTtranspro}, which is given by:

\begin{equation}
    \hat{S_c}=\operatorname{diag}\left(\mathbf{T}_{c} 1_{n_c} \right)^{-1} \mathbf{T}_c S_c,
\end{equation}
where $\operatorname{diag}(\cdot)$ is a diagonal matrix.

After obtaining the adapted augment information $\hat{S_c}$, we unite it with the support sample representations to get the $c$-th class prototype:
\begin{equation}
    P_c = mean(union(S_c, \hat{S_c})).
\end{equation}

\subsection{Training and Testing Phases}
\label{sec:TT}


\noindent\textbf{Training Phase} During the training phase, the probability of query sample $x^Q_i$ belonging to the $c$-th class is computed by a softmax function with the Euclidean distances between its representation $v^Q_i$ and the prototypes:
\begin{equation}
    P(y_c \mid x^Q_i, \mathcal{P})= \frac{exp(-|| v^Q_i - P_c||^2_2)}{\sum_{i=1} exp(-|| v^Q_i - P_i||^2_2)}.
    \label{eq:sim}
\end{equation}

We use the cross-entropy loss function:

\begin{equation}
    \mathcal{L} = \sum_{q=1}^{n} \sum_{c=1}^N y_{qc}logP(y_c \mid x^Q_i, \mathcal{P}),
    \label{eq:loss}
\end{equation}
where $y_{qc} =1$ if $x^Q_i$ belongs to the $c$-th class, otherwise $y_{qc} = 0$, $n$ is the number of query samples. By minimizing $\mathcal{L}$ total with gradient descent methods, all the trainable model parameters can be learned.

\noindent\textbf{Testing Phase} In the testing phase, given an $N$-way $K$-shot task, we generate the corresponding adapted query sample representations and combine them with the original support set as the final support set. Finally, we predict the class label for each query sample $x$ by the Prototypical network,
\begin{equation}
    \widetilde{y}= \mathop{\arg \max}\limits_{k} P(y_c \mid x^Q_i, \mathcal{P}).
    \label{eq6}
\end{equation}

\section{Experiments}
\subsection{Datasets}
Following \citep{Chen_Zhang_Mao_Xue_2022}, we conduct experiments on eight text classification datasets, including four news or review classification datasets: \textbf{HuffPost} \citep{Bao_2019}, \textbf{Amazon} \citep{amazon}, \textbf{Reuters} \citep{Bao_2019}, and \textbf{20News} \citep{20news}. The statistics of the datasets are shown in Table \ref{tb:dataset}, and four intent detection datasets: \textbf{Banking77} \citep{banking}, \textbf{HWU64} \citep{HWU64}, \textbf{Clinic150} \citep{clinc}, and \textbf{Liu57} \citep{HWU64}.

\begin{table}[!h]
\setlength{\tabcolsep}{3pt}
\begin{tabular}{lccc}
\toprule[1pt]
\small \textbf{Dataset}   & \small \textbf{Avg. Length} & \small \textbf{Samples} & \small \textbf{Train / Valid / Test} \\ \hline
HuffPost  & 11.48                     & 36900                        & 20 / 5 / 16                           \\
Amazon    & 143.46                    & 24000                       & 10 / 5 / 9                                \\
Reuters   & 181.41                    & 620                         & 15 / 5 / 11                               \\
20News    & 279.32                    & 18828                        & 8 / 5 / 7                                 \\ \hline
Banking77 & 11.77                     & 13083                        & 25 / 25 / 27                          \\
HWU64     & 6.57                      & 11036                        & 23 / 16 / 25                          \\
Liu57     & 6.66                      & 25478                        & 18 / 18 / 18                        \\
Clinc150  & 8.31                      & 22500                        & 50 / 50 / 50                      \\ 
\bottomrule[1pt]
\end{tabular}
\caption{Dataset statistics.}
\label{tb:dataset}
\end{table}

\noindent \textbf{News or Review Classification Datasets} The HuffPost dataset is a news classification dataset with 36, 900 HuffPost news headlines with 41 classes collected from the year 2012 to 2018. The Amazon dataset is a product review classification dataset including 142.8 million reviews with 24 product categories from the year 1996 to 2014. Following \citep{Bao_2019}, we discard multi-label articles and only use 31 classes, having at least 20 articles. The Reuters dataset is collected from Reuters Newswire in 1987. The 20News dataset is a news classification dataset, which contains 18820 news documents from 20 news groups.

\noindent \textbf{Intent Detection Datasets} The Banking77 dataset is a fine-grained intent classification dataset specific to a single banking domain, including 77 classes. The HWU64 dataset is a multi-domain fine-grained intent classification dataset, which contains 11036 utterances covering 64 intents in 21 domains. The Clinic150 dataset contains 150 intents and 23700 examples across 10 domains. Here we ignore these out-of-scope examples. Liu57 is a highly imbalanced intent classification dataset collected on Amazon Mechanical Turk, which is composed of 25478 user utterances from 54 classes.

\subsection{Baselines}

We compare our proposed LAQDA against several well-established few-shot baseline models, which are briefly described as follows: (1) \textbf{PN} \citep{proto_2017} leverages Euclidean distance to measure query examples against the class vector averaged by support examples. (2) {\textbf{MAML} \citep{MAML_2017} trains a favorable initial point for the base learner by utilizing the meta-learning that learns among tasks. (3) \textbf{IN} \citep{Geng2019} learns a generalized class-wise representation by leveraging a dynamic routing algorithm. (4) \textbf{TPN} \citep{TPN} intends to learn to propagate labels from labeled support samples to unlabeled query samples via episodic training and a specific graph construction, which is a powerful transductive few-shot learning method. (5) \textbf{DS-FSL} \citep{Bao_2019} builds an attention generator to get the representations and classifies samples with a ridge regressor. (6) \textbf{MLADA} \citep{Han__2021} introduces an adversarial domain adaptation network in meta-learning systems. (7) \textbf{P-Tuning} \citep{Luo_2021} extracts discriminative sentence representations from the pre-trained language model BERT guided by label semantics. (8) \textbf{PROTAUG} \citep{Dopierre_Gravier_Logerais_2021} utilizes a short-texts paraphrasing model to generate data augmentation of texts and builds an instance-level unsupervised loss upon the prototypical networks, including two variants: unigram and bigram. (9) \textbf{ContrastNet} \citep{Chen_Zhang_Mao_Xue_2022} introduces instance-level and task-level regularization loss into a contrastive learning model based on BERT representations. (10) \textbf{ProtoVerb} \citep{proto_22} introduces contrastive loss in prompt learning to learn class prototypes from training instances. (11) \textbf{DE} \citep{Liu2023} provides two strategies: Way-DE and Shot-DE to calibrate the data distribution by utilizing the top nearest queries. (12) \textbf{TART} \citep{TART} transforms the class prototypes to per-class fixed reference points in task-adaptive metric spaces.

\begin{table*}[!h]
\setlength{\tabcolsep}{3.5pt}
\centering
\begin{tabular}{lcccccccccc}
\toprule
\multirow{2}{*}{\textbf{Methods}} & \multicolumn{2}{c}{\textbf{HuffPost}}  & \multicolumn{2}{c}{\textbf{Amazon}}    & \multicolumn{2}{c}{\textbf{Reuters}}   & \multicolumn{2}{c}{\textbf{20News}}    & \multicolumn{2}{c}{\textbf{Average}}   \\ \cline{2-11} 
                         & 1-shot        & 5-shot        & 1-shot        & 5-shot        & 1-shot        & 5-shot        & 1-shot        & 5-shot        & 1-shot        & 5-shot        \\ \hline
\small PN (NeurIPS 2017)        & 35.7          & 41.3          & 37.6          & 52.1          & 59.6          & 66.9          & 37.8          & 45.3          & 42.7          & 51.4          \\
\small IN (EMNLP 2019)          & 38.7          & 49.1          & 34.9          & 41.3          & 59.4          & 67.9          & 28.7          & 33.3          & 40.4          & 47.9          \\
\small MAML  (ICML 2017)        & 35.9          & 49.3          & 39.6          & 47.1          & 54.6          & 62.9          & 33.8          & 43.7          & 40.9          & 50.8          \\
\small TPN (ICLR 2019)          & 50.6          & 69.5          & 76.0          & 84.9          & 91.4          & 93.1          & 63.0          & 69.4          & 70.3          & 79.2          \\
\small DS-FSL (ICLR 2020)       & 43.0          & 63.5          & 62.6          & 81.1          & 81.8          & 96.0          & 52.1          & 68.3          & 59.9          & 77.2          \\
\small P-Tuning (ACL 2021)      & 54.5          & 65.8          & 62.2          & 79.1          & 90.0          & \textbf{96.7}         & 56.2          & 77.7          & 65.7          & 79.8          \\
\small MLADA (ACL 2021)         & 45.0            & 64.9          & 68.4          & 86.0            & 82.3          & \textbf{96.7}         & 59.6          & 77.8          & 63.8          & 81.4          \\
\small ContrastNet (AAAI 2022)  & 51.8          & 67.8          & 73.5          & 83.6          & 88.5          & 94.6          & 70.9          & 80.5          & 71.2          & 81.6          \\
\small ProtoVerb (ACL 2022)     & 53.1          & 70.8          & 72.4          & 84.7          & 85.4          & 94.2          & 60.2          & 83.1          & 67.8          & 83.2          \\
\small Shot-DE (AAAI 2023)      & 51.9          & 71.4          & 76.1          & 86.9          & 90.6          & 95.1          & 71.0          & 83.2          & 72.4          & 84.2          \\
\small Way-DE (AAAI 2023)       & 51.9          & 71.7          & 76.1          & 87.4          & 90.6          & 95.2          & 71.0          & 83.2          & 72.4          & 84.4          \\
\small TART (ACL 2023)          & 46.5          & 68.9          & 73.7          & 84.3          & 86.9          & 95.6          & 73.2          & 84.9          & 70.1          & 83.4          \\ \hline
\small LAQDA (QDA / o)                    & 50.5              & 69.8              & 73.7             & 87.4             & 88.4          & 95.2          & 71.1          & 84.2          & 70.9          & 84.2          \\
\small LAQDA (LA / o)                     & \textbf{55.8}     & \textbf{72.4}     & \textbf{79.7}    & \textbf{88.6}    & \textbf{91.9} & 92.6          & \textbf{76.6} & \textbf{85.5} & \textbf{76.0}   & 84.8          \\
\small LAQDA (ours)                      & \textbf{57.0}       & \textbf{72.8}     & \textbf{80.0}      & \textbf{88.6}    & \textbf{92.5} & 95.3 & \textbf{77.4} & \textbf{85.7} & \textbf{76.7} & \textbf{85.6} \\
 \bottomrule[1pt]
\end{tabular}
\caption{The 5-way 1-shot and 5-shot average accuracy on news or review classification datasets. The LAQDA (LA / o) model denotes the LAQDA without using our Label-Adapter, and the LAQDA (QDA / o) denotes the LAQDA without Query-Data-Augmenter.}
\label{tb:news}
\end{table*}

\subsection{Implementation Details}

\textbf{Evaluation Metric} Following \citep{Chen_Zhang_Mao_Xue_2022}, we use accuracy (ACC) to evaluate the performance. All reported results are from 5 different runs, and in each run, the training, validation and testing classes are randomly resampled.

\noindent\textbf{Parameter Setting} We follow \citep{Chen_Zhang_Mao_Xue_2022} to conduct experiments on 5-way 1-shot and 5-shot setting, randomly sample 100, 100, and 1000 tasks for each training, validation, and testing epoch in all the approaches. In the news and review classification task, the number of query samples per class in each episode is 25. In the intent detection task, the number of query samples per class in each episode is 5. In terms of the Word Representation Layer, we use the pure pre-trained \texttt{bert-base-uncased} model for the news or review classification task and use the further pre-trained BERT language model provided in \citep{Dopierre_Gravier_Logerais_2021} for the intent detection task. We set $R = 10$ for the news or review classification task, while $R = 4$ for the intent detection task. We adopt the AdamW \citep{adamw} algorithm with a learning rate of 1e-6 as the optimizer and execute early stopping when the performance of the validation set fails to increase within 20 epochs. Specific settings can be found in our publicly available repository. All the experiments are conducted with NVIDIA RTX A6000 GPUs (20 epochs per hour).

\subsection{Results Analysis}

Tables \ref{tb:news} and \ref{tb:intent} report the experimental results for the news or review classification task and the intent detection task. Some baseline results are taken from \citep{Liu2023, TART} and the top 2 results are highlighted in bold.

\noindent \textbf{News or Review Classification} From Table \ref{tb:news}, we can make the following observation: (1) Our LAQDA achieves the best performance in average. Specifically, LAQDA achieves significant performance improvement over existing methods by 4.3\%-36.3\% and 1.4\%-37.7\% in the 1-shot and 5-shot scenarios, indicating that our model contributes more to a generation of distinguishable class representation, particularly when the number of labeled class samples is small. (2) LAQDA performs much better than the baselines in nearly all the cases (with only one exception). This is because Reuters has similar text characteristics, MLADA and P-tuning can make better use of base class information, but our approaches still outperform them significantly in the 1-shot scenario.

\noindent \textbf{Intent Detection} From Table \ref{tb:intent}, it is easy to find that: (1) Compared with these latest methods, the proposed LAQDA can achieve very competitive performance. Specifically, LAQDA achieves 90\% accuracy across all four datasets. (2) Limited by the number of queries, the improvement of LAQDA is affected, two results were sub-optimal, but only by a few tenths of a percent, which still validates the effectiveness of the proposed strategies.

\noindent \textbf{Analysis} LAQDA achieves a more significant boost in the 1-shot scenario than in the 5-shot scenario. This is because the fewer samples in the support set, the more challenging for other methods to calculate accurate prototypes due to the randomness of the sampled support sets and intra-class differences. LAQDA not only utilizes label names to make the sample representations more suitable in the task-adaptive metric space to get intra-class closer sample representations but also estimates prototypes using query samples, mitigating the inaccuracy of prototypes merely calculated by support samples.

\begin{table*}[!h]
\setlength{\tabcolsep}{3.5pt}
\centering
\begin{tabular}{lcccccccccc}
\toprule
\multirow{2}{*}{\textbf{Methods}} & \multicolumn{2}{c}{\textbf{Banking77}} & \multicolumn{2}{c}{\textbf{HWU64}} & \multicolumn{2}{c}{\textbf{Liu57}} & \multicolumn{2}{c}{\textbf{Clinic150}} & \multicolumn{2}{c}{\textbf{Average}} \\ \cline{2-11}
                                  & 1-shot             & 5-shot            & 1-shot           & 5-shot          & 1-shot           & 5-shot          & 1-shot             & 5-shot            & 1-shot            & 5-shot           \\ \hline
\small PROTAUG (ACL 2021)             & 86.9               & 94.5              & 82.4             & 91.7            & 84.4             & 92.6            & 94.9               & 98.4              & 87.2              & 94.3             \\
\small PROTAUG (bigram)              & 88.1               & 94.7              & 84.1             & 92.1            & 85.3             & 93.2            & 95.8               & 98.5              & 88.3              & 94.6             \\
\small PROTAUG (unigram)             & 89.6               & 94.7              & 84.3             & 92.6            & 86.1             & 93.7            & 96.5               & 98.7              & 89.1              & 94.9             \\
\small ContrastNet (AAAI 2022)            & 91.2               & \textbf{96.4}     & 86.6             & 92.6            & 85.9             & 93.7            & 96.6               & 98.5              & 90.1              & 95.3             \\
\small Shot-DE (AAAI 2023)                & 90.5               & 95.8              & 87.1             & 93.5            & 90.4             & 95.2            & 98.0                 & 99.2              & 91.5              & 95.9             \\
\small Way-DE (AAAI 2023)                 & 90.5               & 95.4              & 87.1             & 93.4            & 90.4             & \textbf{95.5}   & 98.0                 & \textbf{99.3}     & 91.5              & 95.9             \\ \hline
\small LAQDA (QDA / o)                      & 89.8               & 96.0                & 85.7             & 93.6            & 88.6             & 95.2            & 96.8               & 99.0                & 90.2              & 96.0               \\
\small LAQDA (LA / o)                       & \textbf{93.0}        & 96.0                & \textbf{90.1}    & \textbf{93.8}   & \textbf{92.3}    & \textbf{95.7}   & \textbf{98}        & \textbf{99.2}     & \textbf{93.4}     & \textbf{96.2}    \\
\small LAQDA (ours)                       & \textbf{92.5}      & \textbf{96.2}     & \textbf{90.0}      & \textbf{94.0}     & \textbf{92.5}    & 95.3            & \textbf{98.4}      & \textbf{99.2}     & \textbf{93.4}     & \textbf{96.2}   \\
\bottomrule[1pt]
\end{tabular}
\caption{The 5-way 1-shot and 5-shot average accuracy on intent detection datasets.}
\label{tb:intent}
\end{table*}

\begin{table*}[h!]
\setlength{\tabcolsep}{3.5pt}
\centering
\begin{tabular}{lcccccccccc}
\toprule
\multirow{2}{*}{\textbf{Methods}} & \multicolumn{2}{c}{\textbf{HuffPost}}  & \multicolumn{2}{c}{\textbf{Amazon}}    & \multicolumn{2}{c}{\textbf{Reuters}}   & \multicolumn{2}{c}{\textbf{20News}}    & \multicolumn{2}{c}{\textbf{Average}}   \\ \cline{2-11} 
                         & 1-shot        & 5-shot        & 1-shot        & 5-shot        & 1-shot        & 5-shot        & 1-shot        & 5-shot        & 1-shot        & 5-shot        \\ \hline
\small  FastText-PN                         & 31.6              & 53.7              & 46.8             & 67.9             & 56.6              & 76.1             & 34.3             & 47.7             & 42.3              & 61.3             \\
\small  FastText-PN (QDA / w)                   & \textbf{39.5}     & \textbf{54.8}     & \textbf{55.0}      & \textbf{69.0}      & \textbf{70.6}     & \textbf{83}      & \textbf{40.8}    & \textbf{50.7}    & \textbf{51.5}     & \textbf{64.4}   \\ \hline

\small Bert-PN                  & 37.4          & 55.9          & 50.9          & 73.2          & 44.8          & 64.1 & 39.1          & 56.2 & 43.1          & 62.4 \\
\small Bert-PN (QDA / w)          & \textbf{41.6} & \textbf{56.1} & \textbf{60.9} & \textbf{75.7} & \textbf{45.4} & \textbf{64.4 }        & \textbf{43.9} & \textbf{56.5 }         & \textbf{48.0} & \textbf{63.2}          \\ \hline
\small LAQDA (LA / o | ICL)   & 53.0	   &69.4	  &79.8	  &88.0	    &92.4	  &94.1	  &77.4	  &85.5	  &75.6	  &84.3 \\
\small LAQDA (ours)     & \textbf{57.0} & \textbf{72.8} & \textbf{80.0} & \textbf{88.6} &     \textbf{92.5}     & \textbf{95.3}          & \textbf{77.4} & \textbf{85.7} & \textbf{76.7} & \textbf{85.6} \\
\bottomrule[1pt]
\end{tabular}
\caption{The ablation study results on news or review classification datasets. The LAQDA (LA /o | ICL) denotes the LAQDA using In-context learning instead of our Label-Adapter. The Bert-PN and FastText-PN denote adding a Prototype Networks classifier on pure pre-trained \texttt{bert-base-uncased} and FastText using the mean of word vector as sample representations. The (QDA / w) denotes upgraded versions with our QDA module. 
}
\label{tb:ablation}
\end{table*}

\subsection{Ablation Study}

\noindent\textbf{The effectiveness of LA \& QDA} 
From Table \ref{tb:news} and \ref{tb:intent}, we can observe that: (1) With Query-Data-Augmenter,  LAQDA  improves few-shot text classification performance upon LAQDA (QDA / o); (2) LAQDA further promotes LAQDA (LA / o) by adding Label-Adapter. These results demonstrate the effectiveness of our proposed LA and QDA modules, which utilize the information within the task itself to mitigate the overfitting issue caused by a limited number of labeled samples. In the task of intention recognition, the role of LA is not obvious, even 0.3\%, and 0.1\% negative growth on the 1-shot setting of the Banking77 and HWU64 datasets. It is because the sentences of the intent datasets are short (average 10 words), with an additional 5 class representations, it is not conducive to LA further extracting information related to the class.

\noindent\textbf{LA vs In-context Learning} We also try to utilize the PLMs ability of in-context to do Label-Adapter, which adds task label names directly in sentences. From Table \ref{tb:ablation}, it can be seen that LAQDA (LA / o | ICL)'s scores drop by 1.1\% and 1.3\% averagely in the 1-shot and 5-shot scenarios compared to LAQDA. Its scores are lower than the LAQDA (LA / o), which demonstrates the effectiveness of our proposed LA module. This is because the class names may contain redundant tokens, ICL may interfere with sample representation, but LA aligns labels and samples in a high-level semantic space, which effectively reduces intra-class differences.

\noindent\textbf{QDA vs TPN \& DE} To further demonstrate the effectiveness of our proposed QDA module, we compare LAQDA (LA / o) to other methods using the query samples. TPN leverages query samples to construct a graph classifier, and Way-DE leverages query samples to do distribution calibration. From Table \ref{tb:news}, it can be seen that LAQDA (LA / o) has the best scores. Specifically, LAQDA (LA / o) achieves 5.7\% and 5.6\% better than TPN, 3.6\% and 0.4\% better than Way-DE in the 1-shot and 5-shot scenarios. In addition, as shown in Table \ref{tb:ablation}, QDA also plays a role in pure pre-trained Bert-PN and FastText-PN, especially in the 1-shot scenario. This also demonstrates the versatility of the QDA module.

\begin{figure*}[!t]
\centering
\subfigure[Bert-PN]{
\label{orgvisa}
\includegraphics[width=0.3\textwidth]{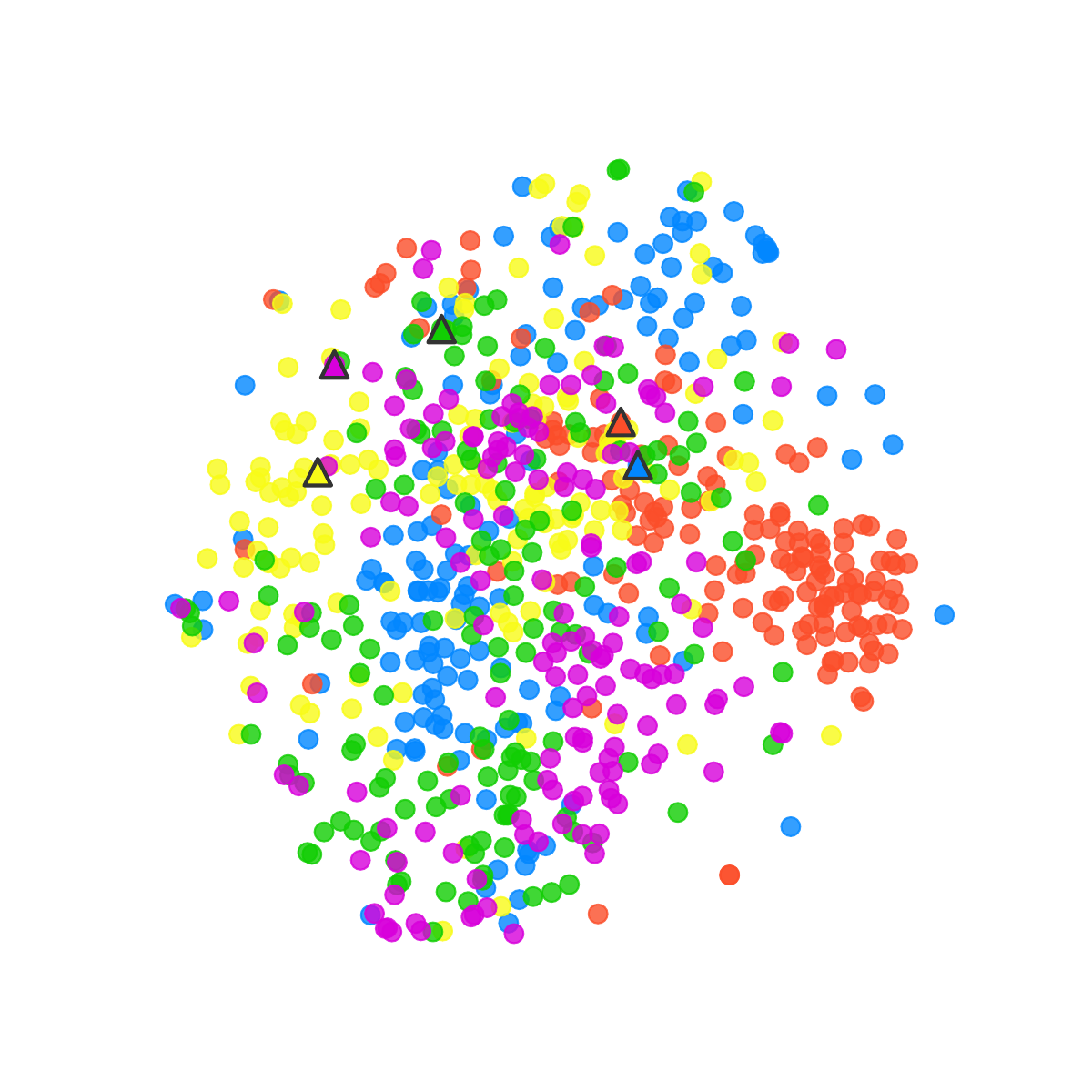}}
\subfigure[LAQDA (QDA /o)]{
\label{orgvisb}
\includegraphics[width=0.3\textwidth]{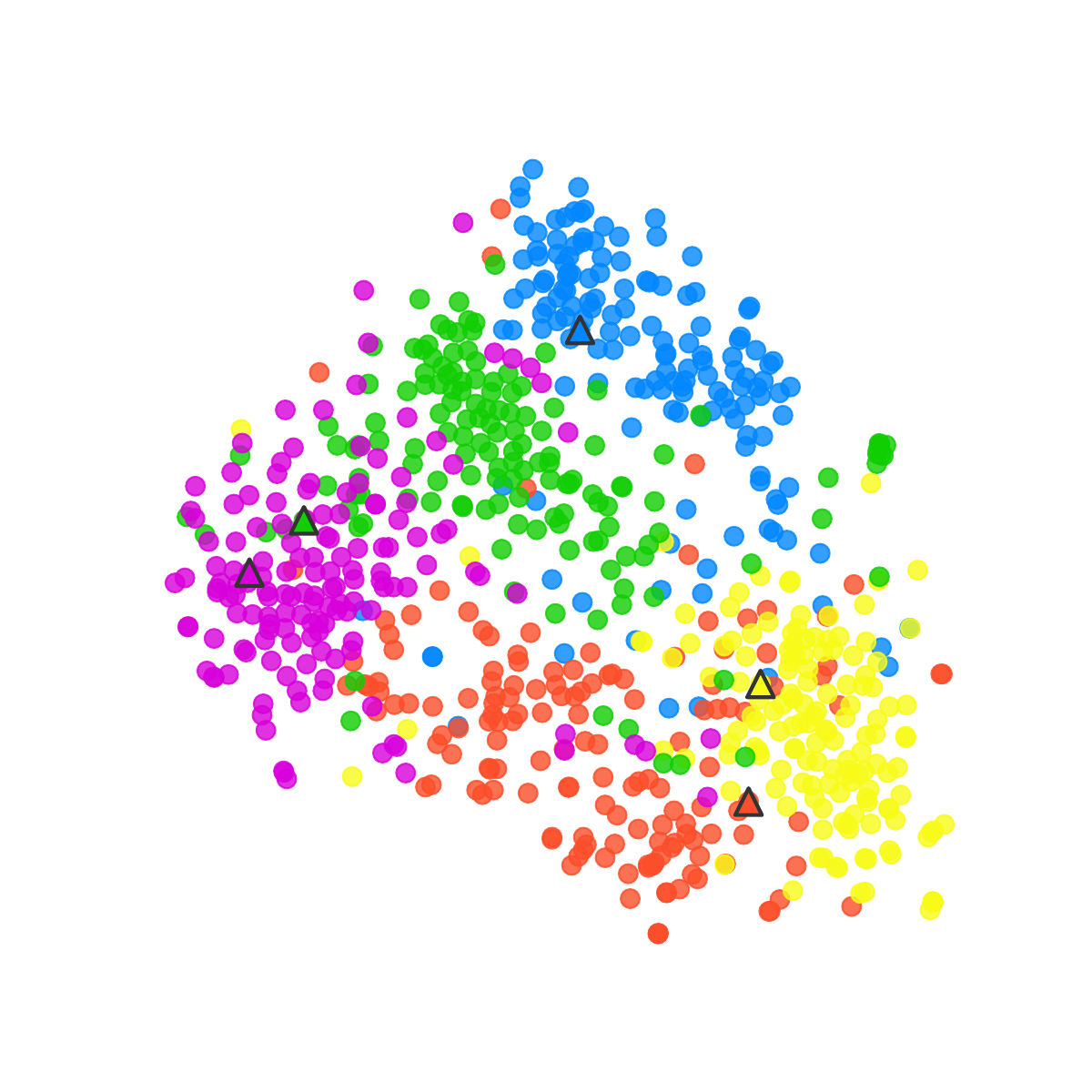}}
\subfigure[LAQDA]{
\label{orgvisc}
\includegraphics[width=0.3\textwidth]{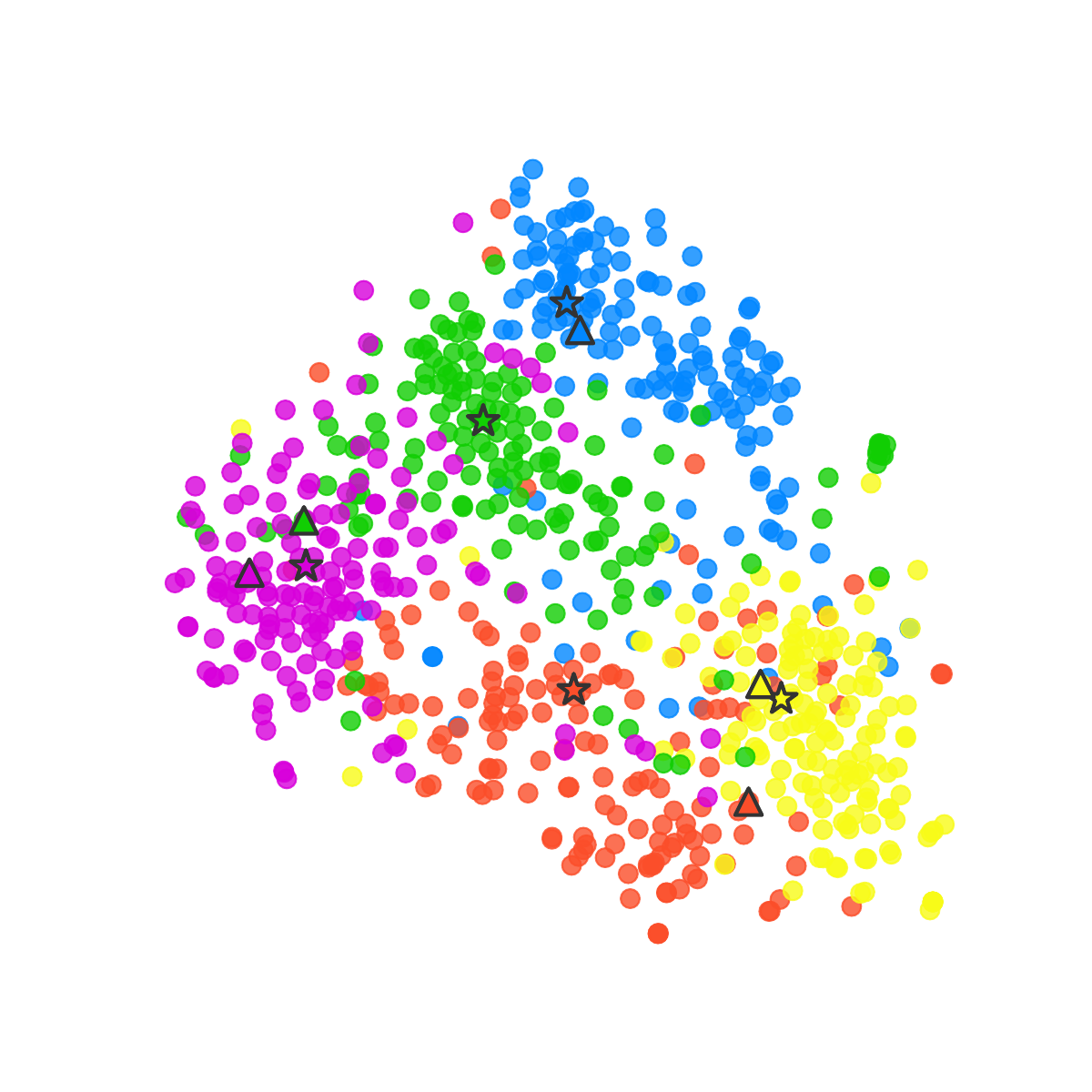}}
\caption{Visualization of sample text representations sampled from five novel classes on HuffPost dataset. The triangles represent the class prototypes calculated by support set and the pentagrams represent the class prototypes calculated with our whole model.}
\label{orgfigvis}
\end{figure*}

\subsection{Visualization}
To investigate models' ability in calculating better prototypes, we visualize the sample representations and class prototypes produced by Bert-PN, LAQDA (QDA /o) and LAQDA using t-SNE~\citep{Maaten2008VisualizingDU}. We randomly select 5 classes on HuffPost dataset, where 5 samples as the support set and 150 samples as the query set per class. The results are shown in Figure~\ref{orgfigvis}.  It can be observed that the text representation generated by LAQDA (QDA /o) in Figure~\ref{orgvisb} is more discriminative than that of the vanilla Bert-PN in Figure~\ref{orgvisa}. Due to the randomness of the sampled support sets and large intra-class differences, the prototypes that are obtained from the support set are not accurate and representative, like green and orange. On the contrary, our proposed LAQDA calculates the prototypes by using the query sample information, which makes prototypes easier to be distinguished and each prototype closer to its class center, as shown in Figure~\ref{orgvisc}.

\section{Related Work}

\subsection{Transfer Learning Based Methods}
Transfer learning aims to tackle few-shot text classification by leveraging knowledge from source domains to target domains. But fine-tuning Pre-trained Language Models (PLMs)~\citep{Devlin_2019, ROBERT_2019} is still suboptimal due to the gap between pre-training and downstream tasks. Prompt learning, inspired by the ``in-context learning'' approach proposed by GPT-3~\citep{gpt_2020}, has recently gained attention for its ability to stimulate model knowledge with just a few prompts, which converts the classification task to a cloze-style mask language modeling problem. Typical prompt learning methods focus on designing a prompt template or expanding the label words to improve the ability of large-scale models in a few labeled sample scenes. PET~\citep{Schick_2021} constructs a prompt learning paradigm for few-shot text classification, which needs designing the template manually. KPT family~\citep{KPT, KPT++} construct an external knowledge graph for PLMs to predict query labels. These methods often require large-scale language models (LLMs) and are more suitable for explicit and simplistic classification tasks such as emotion recognition (positive or negative), may not be applicable in many real-world scenarios.

\subsection{Meta Learning Based Methods}
Meta-learning aims to learn from different small tasks of source classes in the training set and generalizes to unseen tasks of target classes in the test set. Existing methods are mainly divided into two categories: (1) Optimization-based methods learn to find a good initialization parameter to adapt with few-shot training examples, such as MAML~\cite{MAML_2017} and Reptile~\cite{MAML++_2018} attempt to find initial parameters through a few-shot gradient update mechanism. And \textbf{AMGS}~\cite{Lei_Hu_Luo_Peng_Wang_2022} adds the Masked Language Modeling task as an auxiliary task and optimizes meta-learner via gradient similarity between it and the basic task. (2) Metric-based methods learn a metric between samples and classes, such as Matching Networks~\citep{match_2016} with cosine similarity, Prototypical Networks~\citep{proto_2017} with Euclidean distance, Relation Networks~\citep{relation_2017} with convolutional neural networks. (3) Data-augmentation-based methods try to augment the data to help calculate better prototypes. DE~\citep{Liu2023} utilizes the top nearest queries to calibrate the data distribution and generate more informative samples. MEDA~\citep{MEDA} leverage data augmentation to expansion support set. Our approach combines the ideas of measurement and enhancement, the key idea is using label names to generate better sample representations and using highly similar query samples as data extensions by optimal transport.

\section{Conclusion}

We propose a meta-learning method called LAQDA for few-shot text classification, fully utilizing the information within the task to mitigate the overfitting issue caused by a limited number of labeled samples. Specifically, we propose two key modules: \textbf{L}abel-\textbf{A}dapter uses label information to construct a task-adaptive metric space that generates intra-class closer and inter-class differences larger sample representations. \textbf{Q}uery-\textbf{D}ata-\textbf{A}ugmenter leverages the query samples to calculate the class prototypes, mitigating the problem of inaccurate prototypes caused by the randomness of the sampled support sets and intra-class differences. Last but not least, we merely use the information from the task itself instead of introducing external knowledge or LLMs. Extensive experiments are conducted on eight benchmark datasets, and our approach outperforms the state-of-the-art methods.

\section{Limitations}
Our approach focuses on making better use of information from the task itself, such as label names and query samples, which in some scenarios may not be as effective as using an external knowledge base. In addition, our method is primarily suitable for text classification, such as news category or product review classification. It is not appropriate for text generation tasks. Lastly, our approach is based on meta-learning and only 6 layers of \texttt{bert-base-uncased} are fine-tuned, using large and complex feature encoders like LLMs may pose scalability challenges.

\section{Acknowledgments}
This work was supported by National Natural Science Foundation of China (No. 62476040).

\bibliography{anthology,custom}
\bibliographystyle{acl_natbib}

\appendix

\section{Appendix}

\subsection{Pseudocode}
Our method mainly consists of four modules. First, the \textit{Word Representation Layer} gets the word vector representations from the input sentences and label names. Second, the \textit{Label-Adapter} joints the label names and samples to generate the intra-class closer sample representations. Third, the \textit{Query-Data-Augmenter} leveraging query samples as data augment to calculate the class prototypes, mitigating the problem of inaccurate and ambiguous support set samples caused by probable large intra-class differences. Finally, query samples are inferred by a\textit{ Classifier}. The whole training procedure for LAQDA is summarized in Algorithm \ref{alg:algorithm}.

\begin{algorithm}[]
\caption{Training procedure of LAQDA}
\label{alg:algorithm}
\textbf{Input}: Training data $\{X_{train}, Y_{train}\}$; $T$ episodes and $ep$ epochs; $N$ classes in support set or query set; $K$ samples in each class in the support set and $M$ samples in each class in the query set; Word Representation Layer $f_\Phi$; Label-Adapter $LA_\theta$; Query-Data-Augmenter $QDA$. \\
\textbf{Output}: Parameters $\Phi, \theta$ after training \\
\begin{algorithmic}[1] 
\FOR{$i \in [1,ep]$} 

\STATE      $\mathcal{Y} \leftarrow  \Lambda(Y_{train}, N)$;   \textit{ //  select $N$ elements from $Y_{train}$ randomly.}
\FOR {each $j \in [1,T]$}
\STATE      $S, Q, L\leftarrow \emptyset, \emptyset, \emptyset$;
\FOR{$y \in \mathcal{Y}$}
\STATE        $ S \leftarrow S \cup \Lambda(X_{train}\{y\},K)$;
\STATE        $ Q \leftarrow Q \cup \Lambda(X_{train}\{y\} /\ S,M)$;
\STATE        $ L \leftarrow  \Omega (N)$;     \textit{ //  get the task label names }
\ENDFOR
\STATE        $v^S, v^Q \leftarrow LA(f_\Phi(S,Q,L))$ ;
\STATE        $\mathcal{P} \leftarrow QDA(v^S, v^Q)$;
\STATE  Update $\Phi, \theta$ by the loss of the Eq. \ref{eq:loss};
\ENDFOR
\ENDFOR
\end{algorithmic}
\end{algorithm}

\subsection{Experiment Setting}
\subsubsection{Datasets}
Following \citep{Chen_Zhang_Mao_Xue_2022}, we evaluate our method LAQDA under typical 5-way tasks on four news or review classification datasets: HuffPost, Amazon, Reuters, and 20News. Additionally, we follow \citep{Chen_Zhang_Mao_Xue_2022} to evaluate our method on intent detection datasets: Banking77, HWU64, Clinic150, and Liu57. The average length of sentences in news or review classification datasets is much longer than those in intent detection datasets. Table \ref{tb:dataset} concludes the statistics of all datasets. To fully evaluate our approach, we also conduct experiments on RCV1 and FewREL datasets.

\noindent \textbf{Typical News or Review Classification Datasets}

\begin{table*}[!h]
\setlength{\tabcolsep}{3.5pt}
\centering
\begin{tabular}{lcccccccc}
\toprule
\small \textbf{Hyperparameter} & \textbf{Banking77} & \textbf{HWU64} & \textbf{Liu57} & \textbf{Clinic150} & \textbf{HuffPost} & \textbf{Amazon} & \textbf{Reuters} & \textbf{20News} \\ \hline
optimizer               & AdamW              & AdamW          & AdamW          & AdamW              & AdamW             & AdamW           & AdamW            & AdamW           \\
epochs                  & 100                & 100            & 100            & 100                & 100               & 100             & 100              & 100             \\
episodeTrain            & 100                & 100            & 100            & 100                & 100               & 100             & 100              & 100             \\
episodeVal              & 100                & 100            & 100            & 100                & 100               & 100             & 100              & 100             \\
episodeTest             & 1000               & 1000           & 1000           & 1000               & 1000              & 1000            & 1000             & 1000            \\
learning rate           & 1e-6           & 1e-6       & 1e-6       & 1e-6           & 1e-6          & 1e-6        & 1e-6         & 1e-6        \\
warmup steps            & 100                & 100            & 100            & 100                & 100               & 100             & 100              & 100             \\
weight decay            & 0.1                & 0.1            & 0.1            & 0.1                & 0.1               & 0.1             & 0.1              & 0.1             \\
dropout rate            & 0.1                & 0.1            & 0.1            & 0.1                & 0.1               & 0.1             & 0.1              & 0.1             \\
R                       & 4                  & 4              & 4              & 4                  & 10                & 10              & 10               & 10              \\
query per class         & 5                  & 5              & 5              & 5                  & 25                & 25              & 15               & 25              \\
freeze layers          & 6                  & 6              & 6              & 6                  & 6                 & 6               & 6                & 6              \\
\bottomrule[1pt]
\end{tabular}
\label{tb:hp}
\caption{The specific hyperparameters used by each dataset. $R$ is the number of query set samples for the Query-Data-Augmenter.}
\end{table*}

\begin{table*}[!t]
\centering
\begin{tabular}{lcccccc}
\toprule
\multirow{2}{*}{\textbf{Method}} & \multicolumn{2}{c}{\textbf{RCV1}} & \multicolumn{2}{c}{\textbf{FewRel}} & \multicolumn{2}{c}{\textbf{Average}} \\ \cline{2-7}
                                 & 1-shot          & 5-shot          & 1-shot           & 5-shot           & 1-shot            & 5-shot           \\ \hline
DS-FSL (ICLR 2020)                & 54.1            & 75.3            & 67.1             & 83.5             & 60.6              & 79.4             \\
ContrastNet (AAAI 2022)           & 65.7            & 87.4            & 85.3             & 92.7             & 75.5              & 90.1             \\
TART (ACL 2023)                  & 65.3            & 81.1            & 83.5             & 92.6             & 74.4              & 86.9             \\ \hline
LAQDA (ours)                      & \textbf{77.2}   & \textbf{87.0}   & \textbf{92.6}    & \textbf{95.1}    & \textbf{84.9}     & \textbf{91.1}   \\
\bottomrule[1pt]
\end{tabular}
\caption{The 5-way 1-shot and 5-shot average accuracy on RCV1 and FewRel datasets.}
\label{tb:other}
\end{table*}
\textbf{HuffPost} \citep{Bao_2019} consists of news headlines published on HuffPost between 2012 and 2018. These headlines are split into 41 classes. In addition, their sentences are shorter and less grammatically correct than formal phrases.

\textbf{Amazon}~\citep{amazon} consists of 142.8 million customer reviews from 24 product categories. Following~\citep{Han__2021}, we use a subset with 1000 reviews per category.

\textbf{Reuters}~\citep{Bao_2019} consists of shorter Reuters articles in 1987. Following~\citep{Bao_2019}, we discard multi-label articles and use 31 classes, each with at least 20 articles.

\textbf{20News} \citep{20news} is a collection of approximately 20,000 newsgroup documents, partitioned equally among 20 different newsgroups.

\noindent \textbf{Typical Intent Detection Datasets}

\textbf{Banking77} \citep{banking} is a fine-grained single-domain dataset for intent detection, in which some categories are similar and may have overlap with others.

\textbf{HWU64} \citep{HWU64} contains 11036 utterances covering 64 intents in 21 domains. The examples are from a real-world home robot, with multi-domain utterances, e.g., email, music, weather and so on.

\textbf{Liu57} \citep{HWU64} is collected from Amazon Mechanical Turk, which is composed of 25478 user utterances from 54 classes.

\textbf{Clinic150} \citep{clinc} contains 150 intents and 23700 examples across 10 domains. It has 22500 user utterances evenly distributed in every intent and 1200 out-of-scope queries. Here we ignore these out-of-scope examples.

\noindent \textbf{Another Less Used Datasets}

\textbf{FewRel} \citep{FewREL} is a relation classification dataset developed for few-shot learning. Each example is a single sentence, annotated with a head entity, a tail entity, and their relation. The goal is to predict the correct relation between the head and tail. The public dataset contains 80 relation types.

\textbf{RCV1} \citep{RCV1} is a collection of Reuters newswire articles from 1996 to 1997. These articles are written in formal speech and labeled with a set of topic codes. We consider 71 second-level topics as our total class set and discard articles that belong to more than one class.

\subsubsection{Evaluation Metric}
Following \citep{Chen_Zhang_Mao_Xue_2022}, we use accuracy (ACC) to evaluate the performance. Because the setting of N-way K-shot is class-balanced, it makes sense to use ACC only. To test the stability of our method, we perform a five-fold class split for each dataset, following the approach outlined in \citep{Chen_Zhang_Mao_Xue_2022}. All reported results are from 5 different runs, and in each run, the training, validation, and testing classes are randomly resampled.

\subsubsection{Parameter Setting} We follow \citep{Chen_Zhang_Mao_Xue_2022} to conduct experiments on the 5-way 1-shot and 5-shot setting, randomly sample 100, 100, and 1000 tasks for each training, validation, and testing epoch in all the approaches. In news and review classification task, the number of query samples per class in each episode is 25. In intent detection task, the number of query samples per class in each episode is 5. In terms of the Word Representation Layer, we use the pure pre-trained \texttt{bert-base-uncased} model for the news or review classification task and use the further pre-trained BERT language model provided in \citep{Dopierre_Gravier_Logerais_2021} for the intent detection task. To save computing resources, we only fine-tune the last 6 layers of BERT parameters. We set $R = 10$ for the news or review classification task, while $R = 4$ for the intent detection task. We adopt the AdamW\citep{adamw} algorithm with a learning rate of 1e-6 as the optimizer and execute early stopping when the performance of the validation set fails to increase within 20 epochs. All the experiments are conducted with NVIDIA RTX A6000 GPUs (20 epochs per hour). The specific parameters are shown in the Table \ref{tb:hp}. It is easy to find that our method maintains the same set of hyperparameters on different datasets, which indicates that our method is general. Except for the Reuters dataset, the number of query samples was adjusted to 15 because there were only 20 samples per class. We use the same parameters as Liu57 on RCV1 and FewRel datasets.

\subsubsection{More Results}

Since baselines do not conduct experiments on the two datasets RCV1 and FewRel and DE(AAAI 2023) does not disclose the code. We conducted five runs of the latest method TART (ACL 2023) and our method respectively, and the experimental results are shown in Table \ref{tb:other}. It can be seen that the effectiveness of our method is significantly improved compared with TART (ACL 2023).

\end{document}